\pdfoutput=1

\documentclass[11pt]{article}

\usepackage{EMNLP2023}

\usepackage{times}
\usepackage{latexsym}

\usepackage[T1]{fontenc}

\usepackage[utf8]{inputenc}

\usepackage{microtype}

\usepackage{inconsolata}

\usepackage{subfigure}
\usepackage{xspace}
\usepackage{graphicx}
\usepackage{textcomp}
\usepackage{booktabs}
\usepackage{amsmath}
\usepackage{multirow}
\usepackage{multicol}
\usepackage[ruled,vlined]{algorithm2e}

\def\figref#1{Figure~\ref{fig:#1}}
\def\figlabel#1{\label{fig:#1}\label{p:#1}}

\def\tabref#1{Table~\ref{tab:#1}}

\def\tablabel#1{\label{tab:#1}\label{p:#1}}

\title{Unleashing the Multilingual Encoder Potential: \\Boosting Zero-Shot Performance via Probability Calibration}

\author{Ercong Nie$^{1,2}$ \qquad Helmut Schmid$^{1}$ \qquad Hinrich Sch\"utze$^{1,2}$ \\
$^{1}$Center for Information and Language Processing (CIS), LMU Munich, Germany \\
$^{2}$ Munich Center for Machine Learning (MCML), Germany \\
\texttt{nie@cis.lmu.de}}

\begin{document}
\maketitle
\begin{abstract}
 Pretrained multilingual encoder models can directly perform zero-shot multilingual tasks or linguistic probing by reformulating the input examples into cloze-style prompts.
 This is accomplished by predicting the probabilities of the label words at the masked token position, without requiring any updates to the model parameters.
 However, the performance of this method is limited by the model's bias toward predicting label words which frequently occurred during the pretraining. 
 These words typically receive high probabilities. 
 To address this issue, we combine the models with \emph{calibration} techniques which modify the probabilities of label words predicted by the models.
We first validate the effectiveness of a proposed simple calibration method together with other existing techniques on monolingual encoders in both zero- and few-shot scenarios.
We subsequently employ these calibration techniques on multilingual encoders, resulting in substantial performance improvements across a wide range of tasks\footnote{The code and data for this work are publicly available: \url{https://github.com/ercong21/calibration}.}.
\end{abstract}

\section{Introduction}

Prompt-based learning~\citep{brown2020language, liu2021pre} has emerged as an important research area.
Recent research demonstrates that multilingual encoder models are capable of accomplishing zero-shot cross-lingual learning~\citep{zhao-schutze-2021-discrete, huang-etal-2022-zero} and linguistic probing~\citep{shapiro-etal-2021-multilabel-approach, hartmann-etal-2021-multilingual} by using cloze-style prompts. These prompts consist of an input sample, a task-specific context and a \texttt{mask} token.
The encoder model applies Masked Language Modeling (MLM)~\citep{devlin-etal-2019-bert} to generate predictions for the \texttt{mask} token using a selection of prescribed candidate tokens from the vocabulary. These predictions can be subsequently utilized for label classification or probing purposes.
For example, the sentiment analysis of assigning the product review ``\textit{Worked as stated!}'' to class \textsc{pos} can be reformulated as: ``\textit{\underline{Worked as stated! }} \texttt{All in all, it was [MASK]}.'' The model is requested to fill in the word ``\textit{good}'' at the \texttt{mask} token position, which is mapped to the \textsc{pos} label.

However, earlier studies indicate that the output of masked token prediction is biased towards certain label words in the candidate token list~\citep{weissweiler-etal-2022-better, nie-etal-2023-crosslingualra}. 
This bias not only affects the predicted class probabilities~\citep{holtzman-etal-2021-surface, ahuja-etal-2022-calibration}, but also deteriorates the model's overall performance~\citep{pmlr-v139-zhao21c, lu-etal-2022-fantastically}. According to~\citet{weissweiler-etal-2022-better} and~\citet{pmlr-v139-zhao21c}, label words with higher frequency in the pretraining corpus tend to be predicted with higher probabilities. Besides, the prompt context can also influence the degree of bias present in the masked token prediction.

\begin{figure}[h]
    \centering
    \includegraphics[width=.7\linewidth]{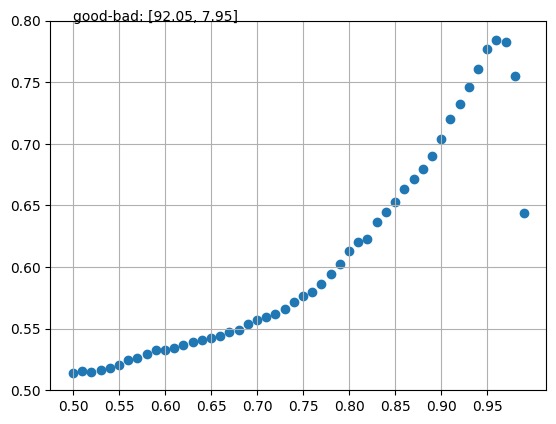}
    \footnotesize
    \caption{Example of the model predictions bias. The graph shows the accuracy on the amazon polarity test data (equally distributed) as a function of the classification threshold. $x$-axis refers to the threshold probability of \texttt{good} to classify examples with the class \textsc{POS}. The best results are obtained by classifying examples as \textsc{pos} if the probability of \texttt{good} exceeds 0.96.}
    \figlabel{influence}
\end{figure}

\begin{table*}[th]
    \centering
    \begin{tabular}{c|l|l|l}
        \toprule
        \textbf{Method} & \textbf{Description} & \textbf{Probability Calculation} &
        \textbf{Source} \\
        \midrule
         CC & {\footnotesize Contextual Calibration} & $\tilde{q}(\textbf{y}|x,t)=\textbf{W}p(\textbf{y}|x,t)+\textbf{b}$ & {\footnotesize \citet{pmlr-v139-zhao21c}}\\
         $\text{PMI}_{\text{DC}}$ & {\scriptsize Domain Conditional Pointwise Mutual Information} & $\tilde{q}(\textbf{y}|x,t)=log\frac{p(\textbf{y}|x,t)}{p(\textbf{y}|t)}$ & {\footnotesize \citet{holtzman-etal-2021-surface}} \\
         CBM & {\footnotesize Calibration By Marginalization} & $\tilde{q}(\textbf{y}|x,t)=\frac{p(\textbf{y}|x,t)}{\frac{1}{|X|}\sum_{x'\in X}p(\textbf{y}|x',t)}$ & {\footnotesize \citet{yang2023improving}} \\
         \hline
         Penalty & {\footnotesize Probability Penalty} & $\tilde{q}(\textbf{y}|x,t)=p(\textbf{y}|x,t)+\textbf{p}$ & {\footnotesize Our proposed method} \\
         \bottomrule
    \end{tabular}
    \caption{Overview of Calibration Methods. \textbf{y} refers to the label words. $X$ is the test dataset, $x$ is an input sample, and $t$ is the prompt template.}
    \tablabel{calibration}
\end{table*}

\figref{influence} illustrates the impact of the above mentioned biases on the model predictions. It shows the results of a binary sentiment analysis task with the $\texttt{BERT}_{\texttt{Base}}$~\citep{devlin-etal-2019-bert} model.
The prompt template and label words used for this example can be found in \tabref{prompts}.
By shifting the threshold for predicting \textsc{pos} from 0.5 to approx.\ 0.95, the performance can be improved by more than 25\%.
Given only a \texttt{mask} token as input, the model predicts 0.92 and 0.08 as probabilities for the label words \texttt{good} and \texttt{bad}, respectively. 
To tackle the bias in the distribution of label words, our proposed solution in this work is to combine pretrained encoder models with \emph{calibration} methods.

In this paper, we contribute by (1) proposing a simple yet effective calibration method that involves adding trainable penalties to the probabilities of the label words, 
(2) demonstrating its effectiveness in achieving performance enhancements comparable to other existing calibration techniques, (3) refining the calibration parameters with only a few training examples for further improvement, and (4) boosting the zero-shot performance of multilingual encoders by introducing calibration methods.

\section{Calibration Methods}

\subsection{Existing Calibration Methods}
\paragraph{Contextual Calibration (CC)}~\citet{pmlr-v139-zhao21c} apply an affine transformation~\citep{platt1999probabilistic} to the original probabilities, as the first equation in \tabref{calibration} shows. The parameters of the affine transformation are obtained from the output probability distribution of the content-free input, e.g., the \texttt{mask} token, denoted $\hat{\textbf{p}}_{cf}$. $\textbf{W}=\text{diag}(\hat{\textbf{p}}_{cf})^{-1}$ is the inverse diagonal matrix of $\hat{\textbf{p}}_{cf}$ and $\textbf{b}$ is an all-zero vector.

\paragraph{Domain Conditional Pointwise Mutual Information ($\text{PMI}_{\text{DC}}$)} \citet{holtzman-etal-2021-surface} adjust the conditional class probability $p(\textbf{y}|x, t)$ by dividing it with the prior probability $p(\textbf{y}|t)$ of that class. We estimate $p(\textbf{y}|t)$ for a given template $t$ using MLM with a prompt created by instantiating the prompt template with an empty input.

\paragraph{Calibration By Marginalization (CBM)}~\citet{yang2023improving} are inspired by $\text{PMI}_{\text{DC}}$. Unlike $\text{PMI}_{\text{DC}}$, CBM approximates $p(\textbf{y}|x,t)$ in a more precise manner by computing its marginalized probability, as the third equation in \tabref{calibration} shows. For each prediction, the sum probability $\Sigma_{x'\in X}p(\textbf{y}|x',t)$ are calculated by taking all test inputs into account.

\subsection{Our Method: Probability Penalty}
Motivated by the observation in \figref{influence} that a simple shift in the model's output distribution can substantially alleviate the label bias, we propose a penalty-based calibration approach as the equation in the last row of \tabref{calibration} shows. The core idea is to introduce a penalty term that is added to each individual label word probability. We initialize the corresponding parameter vector $\textbf{p}$ with the negative prior probabilities of the label words. We estimate these prior probabilities using the output distribution of MLM applied to a \texttt{mask} token as input.

\begin{table*}[t]
\scriptsize
\centering
\begin{tabular}{l|c|c|c|c|c|c|c|c|c|c|c|c} 
               & \multicolumn{5}{c|}{\textbf{Balanced datasets (Acc.)}}                                                                                                                   & \multicolumn{6}{c|}{\textbf{Imbalanced datasets (F1 Score)}}                                                                                                                                    & \multirow{2}{*}{\textbf{Avg.}}      \\ 
\cline{2-12}
               & \multicolumn{1}{l|}{\textbf{AG News}} & \multicolumn{1}{l|}{\textbf{Amazon-P}} & \multicolumn{1}{l|}{\textbf{Amazon-S}} & \multicolumn{1}{l|}{\textbf{XNLI}} & \multicolumn{1}{l|}{\textbf{Yahoo}} & \multicolumn{1}{l|}{\textbf{Pawsx}} & \multicolumn{1}{l|}{\textbf{CoLA}} & \multicolumn{1}{l|}{\textbf{MRPC}} & \multicolumn{1}{l|}{\textbf{QQP}} & \multicolumn{1}{l|}{\textbf{RTE}} & \multicolumn{1}{l|}{\textbf{WNLI}} &                            \\ 
\hline
$\texttt{BERT}_{\texttt{Base}}$     & \multicolumn{1}{l|}{}        & \multicolumn{1}{l|}{}                & \multicolumn{1}{l|}{}            & \multicolumn{1}{l|}{}     & \multicolumn{1}{l|}{}      & \multicolumn{1}{l|}{}         & \multicolumn{1}{l|}{}        & \multicolumn{1}{l|}{}        & \multicolumn{1}{l|}{}       & \multicolumn{1}{l|}{}       & \multicolumn{1}{l|}{}        & \multicolumn{1}{l}{}      \\ 

\hspace{1em} + \textit{no calib.} & 60.2                         & 54.6                                 & 24.8                             & 41.3                      & 36.0                       & 31.2                          & 41.2                         & 46.1                         & 26.9                        & 39.5                        & 29.0                         & 39.2                       \\ 

\hspace{1em} + \textit{CC}             & \textbf{74.6}                         & 61.7                                 & 27.4                             & 41.4                      & 36.2                       & 31.6                          & 51.1                         & 46.1                         & 26.9                        & 39.5                        & \textbf{43.1}                         & \multicolumn{1}{c}{43.6}  \\ 

\hspace{1em} + $\textit{PMI}_{\textit{DC}}$  & 62.1 & 70.8 & 29.9 & 37.9 & 32.1 &	33.8 &	\textbf{51.3} &	44.3 &	49.5 &	38.2 &	30.4 &	43.7     \\ 

\hspace{1em} + \textit{CBM}      &      73.6 &	\textbf{71.3} &	\textbf{33.6} &	\textbf{42.9} &	\textbf{45.2} &	\textbf{49.3} &	49.9 &	\textbf{50.6} &	\textbf{52.6} &	\textbf{50.9} &	42.3 &	\textbf{51.1}         \\ 

\hspace{1em} + \textit{Penalty}        & 67.9                         & 61.7                                 & 26.3                             & 42.6                      & 39.4                       & 31.6                          & 51.1                         & 46.1                         & 26.9                        & 39.5                        & \textbf{43.1}                         & \multicolumn{1}{c}{43.3}  \\ 
\hline
$\texttt{RoBERTa}_{\texttt{Base}}$  & \multicolumn{1}{l|}{}        & \multicolumn{1}{l|}{}                & \multicolumn{1}{l|}{}            & \multicolumn{1}{l|}{}     & \multicolumn{1}{l|}{}      & \multicolumn{1}{l|}{}         & \multicolumn{1}{l|}{}        & \multicolumn{1}{l|}{}        & \multicolumn{1}{l|}{}       & \multicolumn{1}{l|}{}       & \multicolumn{1}{l|}{}        & \multicolumn{1}{l}{}      \\ 

\hspace{1em} + \textit{no calib.} & 76.2                         & 66.1                                 & 24.3                             & 44.0                      & 32.4                       & 31.2                          & 39.6                         & 45.3                         & 26.9                        & 37.1                        & 31.6                         & 41.3                       \\ 

\hspace{1em} + \textit{CC }            & 74.1                         & \textbf{79.5}                                 & 20.0                             & 39.8                      & 15.2                       & 33.7                          & 23.6                         & 46.6                         & 39.8                        & 35.9                        & 32.1                         & \multicolumn{1}{c}{40.0}  \\ 

\hspace{1em} + $\textit{PMI}_{\textit{DC}}$   &    62.3 &	79.4 &	\textbf{34.2} &	45.6 &	25.3 &	43.3 &	43.3 &	\textbf{49.4} &	27.1 &	37.0 &	30.4 &	43.4           \\ 

\hspace{1em} + \textit{CBM}  &\textbf{78.4} &	76.5 &	34.1 &	\textbf{46.4} &	\textbf{42.9} &	\textbf{44.4} &	\textbf{48.2} &	47.5 &	\textbf{50.1} &	\textbf{43.3} &	\textbf{49.0} &	\textbf{51.0}  \\ 

\hspace{1em} + \textit{Penalty}        & 75.6                         & \textbf{79.5}                                 & 30.1                             & 41.4                      & 26.9                       & 33.7                          & 23.6                         & 46.6                         & 39.8                        & 35.9                        & 32.1                         & \multicolumn{1}{c}{42.3}  \\
\hline

\end{tabular}
\caption{Results of zero-shot calibration methods on monolingual tasks. Amazon-P refers to Amazon Polarity (binary classification). Amazon-S refers to Amazon Star (5-way classification).}
\tablabel{results_mono}
\end{table*}

\begin{table*}[!t]
\centering
\footnotesize

\begin{tabular}{l|c|c|c|c|c|c|c|c|c|c|c}
\toprule
\multicolumn{12}{l}{$\texttt{BERT}_{\texttt{Base}}$} \\
\hline
 \multicolumn{2}{c|}{}                        & \multicolumn{2}{c|}{\textbf{AG News}}       & \multicolumn{2}{c|}{\textbf{Amazon-P}}                          & \multicolumn{2}{c|}{\textbf{Pawsx}}                         & \multicolumn{2}{c|}{\textbf{XNLI}}                              & \multicolumn{2}{c}{\textbf{Avg}}  \\
\hline

\multicolumn{2}{c|}{nli-based ZR} & \multicolumn{2}{c|}{$54.9$} & \multicolumn{2}{c|}{$\textbf{82.3}$} & \multicolumn{2}{c|}{$48.2$} & \multicolumn{2}{c|}{$34.8$} & \multicolumn{2}{c}{$55.1$} \\
\hline
                \multicolumn{2}{c|}{calibration}                        & Penalty         & CC                         & Penalty         & CC                         & Penalty         & CC                        & Penalty         & CC                        & Penalty         & CC \\ 
\hline
zero-shot                 & 0                                            & $67.9$      & \multicolumn{1}{c|}{$74.6$} & \multicolumn{1}{c|}{$61.7$} & \multicolumn{1}{c|}{$61.7$} & \multicolumn{1}{c|}{$45.4$} & \multicolumn{1}{c|}{$45.4$} & \multicolumn{1}{c|}{$42.6$} & \multicolumn{1}{c|}{$41.4$} & $54.4$      & $55.8$          \\ 
\hline
\multirow{5}{*}{few-shot} & 1                                            & $65.6_{3.8}$ & $75.7_{1.0}$                 & $67.8_{7.6}$                 & $71.0_{5.6}$             & $51.1_{0.9 }$                & $\underline{51.4_{0.9}}$                 & $42.0_{1.8 }$                & $41.2_{1.9}$                          & $56.6_{3.5}$ & $59.8_{2.4}$     \\ 

                          & 2                                            & $67.2_{3.1}$ & $75.9_{1.6}$                 & $71.9_{4.4}$                 & $\underline{72.2_{3.2}}$                 &  $51.0_{1.1}$                 & $50.7_{1.0}$                 & $42.7_{0.6}$                 & $42.5_{0.9}$                 & $58.2_{2.3}$ & $\underline{\textbf{60.3}_{\textbf{1.7}}}$    \\ 

                          & 4                                            & $67.9_{3.9}$ & $76.6_{0.7}$                 & $73.4_{3.8}$                 & $70.3_{2.9}$                 & $\underline{\textbf{51.6}_{\textbf{1.3}}}$                 & $50.9_{1.3}$                 & $42.8_{0.6}$                 & $\underline{42.8_{0.3}}$                & $58.9_{2.4}$ & $60.2_{1.3}$     \\ 

                      & 8                                            & $69.1_{1.5}$ & $\underline{76.9_{0.1}}$                 & $75.2_{2.3}$                 & $71.8_{1.2}$                & $\underline{\textbf{51.6}_{\textbf{1.1}}}$                & $49.9_{0.6}$                 & $\underline{\textbf{42.9}_{\textbf{0.2}}}$                 & $42.7_{0.2}$                 & $59.7_{1.3}$ & $\underline{\textbf{60.3}_{\textbf{0.5}}}$    \\ 

                      & \multicolumn{1}{r|}{16} & $\underline{69.6_{1.7}}$ & $\underline{\textbf{76.9}_{\textbf{0.1}}}$                 & $\underline{76.0_{1.0}}$                 & $71.4_{1.2}$                 & $51.4_{1.1 }$                & $49.7_{1.0}$                 & $42.8_{0.3}$                 & $42.6_{0.2}$              & $\underline{60.0_{1.0}}$ & $60.2_{0.6}$     \\ 
\midrule
\multicolumn{12}{l}{$\texttt{RoBERTa}_{\texttt{Base}}$} \\
\hline
 \multicolumn{2}{c|}{}                        & \multicolumn{2}{c|}{\textbf{AG News}}       & \multicolumn{2}{c|}{\textbf{Amazon-P}}                          & \multicolumn{2}{c|}{\textbf{Pawsx}}                         & \multicolumn{2}{c|}{\textbf{XNLI}}                              & \multicolumn{2}{c}{\textbf{Avg}}  \\

\hline

\multicolumn{2}{c|}{nli-based ZR} & \multicolumn{2}{c|}{$67.9$} & \multicolumn{2}{c|}{$84.8$} & \multicolumn{2}{c|}{$45.3$} & \multicolumn{2}{c|}{$34.3$} & \multicolumn{2}{c}{$58.1$} \\
\hline
                \multicolumn{2}{c|}{calibration}                        & Penalty         & CC                         & Penalty         & CC                         & Penalty         & CC                        & Penalty         & CC                        & Penalty         & CC \\ 
\hline
zero-shot                 & 0                                            & $75.6$      & \multicolumn{1}{c|}{$74.1$} & \multicolumn{1}{c|}{$79.5$} & \multicolumn{1}{c|}{$79.5$} & \multicolumn{1}{c|}{$45.4$} & \multicolumn{1}{c|}{$45.4$} & \multicolumn{1}{c|}{$41.4$} & \multicolumn{1}{c|}{$39.8$} & $60.5$      & $59.7$          \\ 

\hline

\multirow{5}{*}{few-shot} & 1                                            & $75.6_{2.6}$ & $77.2_{1.5}$                 & $77.4_{8.0}$                 & $\underline{81.3_{4.9}}$                  & $48.4_{1.8}$                 & $48.4_{1.4}$                 & $45.9_{0.9}$                 &$44.8_{1.5}$                 & $61.8_{3.3}$ & $62.9_{2.3}$     \\ 

                          & 2                                            & $73.9_{2.8}$ & $77.3_{1.2}$                 & $81.6_{4.3}$                 & $80.8_{2.4}$                 & $49.0_{1.6}$                 & $48.3_{0.9}$                 & $46.3_{0.7}$                 & $45.8_{0.7}$                  & $62.7_{2.4}$ &
                          $\underline{63.1_{1.3}}$     \\ 

                          & 4                                            & $74.5_{1.9}$ & $77.6_{1.0}$                 & $82.2_{4.4}$                 & $79.6_{1.6}$                 & $49.3_{0.6}$                 & $\underline{48.5_{0.9}}$                 & $\underline{\textbf{47.2}_{0.2}}$                 & $\underline{46.0_{0.3}}$                                 & $63.3_{1.8}$ & $62.9_{1.0}$     \\ 

                          & 8                                            & $76.6_{1.1}$ & $78.1_{0.5}$                 & $\underline{\textbf{85.2}_{\textbf{1.0}}}$                 & $79.7_{1.5}$                 &  $\underline{\textbf{49.6}_{\textbf{0.4}}}$                 & $48.1_{0.7}$                 & $47.1_{0.3}$                 & $\underline{46.0_{1.0}}$                  & $64.6_{0.7}$ & $63.0_{0.9}$     \\ 

                          & \multicolumn{1}{r|}{16} & $\underline{78.3_{0.5}}$ & $\underline{\textbf{78.4}_{\textbf{0.3}}}$                 & $85.1_{1.0}$                 & $79.7_{1.6}$                 & $49.4_{0.6}$                 & $48.1_{0.4}$                 & $47.0_{0.2}$                 & $\underline{46.0_{0.9}}$                 & $\underline{\textbf{65.0}_{\textbf{0.6}}}$ & $\underline{63.1_{0.8}}$     \\
\bottomrule

\end{tabular}
\caption{Results of few-shot calibration methods on monolingual tasks. \textit{nli-based ZR} refers to the NLI-based zero-shot classification baseline~\citep{yin-etal-2019-benchmarking}.}
\tablabel{few_shot_results}
\end{table*}

\section{Experimental Setup}

\paragraph{Dataset} We first validate the effectiveness of the different calibration methods on several monolingual tasks. We study sentiment analysis using two datasets: binary \textbf{Amazon Polarity}~\citep{mcauley2013hidden} and the English subset of 5-label \textbf{Multilingual Amazon Reviews}~\citep{keung-etal-2020-multilingual}, topic categorization using two datasets: the \textbf{Ag News} and \textbf{Yahoo Answers Topics}~\citep{zhang2015character}, sentence pair classification using two datasets: English subsets of \textbf{MNLI}~\cite{conneau-etal-2018-xnli} and \textbf{PAWS-X}~\citep{yang-etal-2019-paws}, and 5 datasets from the GLUE benchmark~\citep{wang2019glue}: \textbf{CoLA}~\citep{warstadt-etal-2019-neural}, \textbf{MRPC}~\citep{dolan2005automatically}, \textbf{QQP}, \textbf{RTE}~\citep{dagan2005pascal}, and \textbf{WNLI}~\citep{levesque2012winograd}. For the evaluation of multilingual encoders, we use \textbf{Multilingual Amazon Reviews}, \textbf{XNLI} and \textbf{PAWS-X}. Besides, following~\citet{nie-etal-2023-crosslingualra}, we expand the \textbf{AG News} dataset to 25 languages using machine translation to conduct a wide range of cross-lingual analyses.

\paragraph{Setup}
In our monolingual experiments, we use the pretrained models \texttt{bert-base-cased}~\citep{devlin-etal-2019-bert} and \texttt{roberta-base}~\citep{liu2019roberta}. In the multilingual experiments, we use their multilingual counterparts \texttt{bert-base-multilingual-cased} and \texttt{xlm-roberta-base}~\citep{conneau-etal-2020-unsupervised}. We use PyTorch~\citep{paszke2019pytorch} and the HuggingFace framework~\citep{wolf2020transformers}. We repeat each experiment 5 times with different random seeds and report the mean and variance. Details of the experimental setting can be found in Appendix \ref{details}.

\section{Results and Analysis}

\begin{figure}[!t]
\subfigure[AG News]{
\minipage{.5\textwidth}
  \includegraphics[width=.95\linewidth]{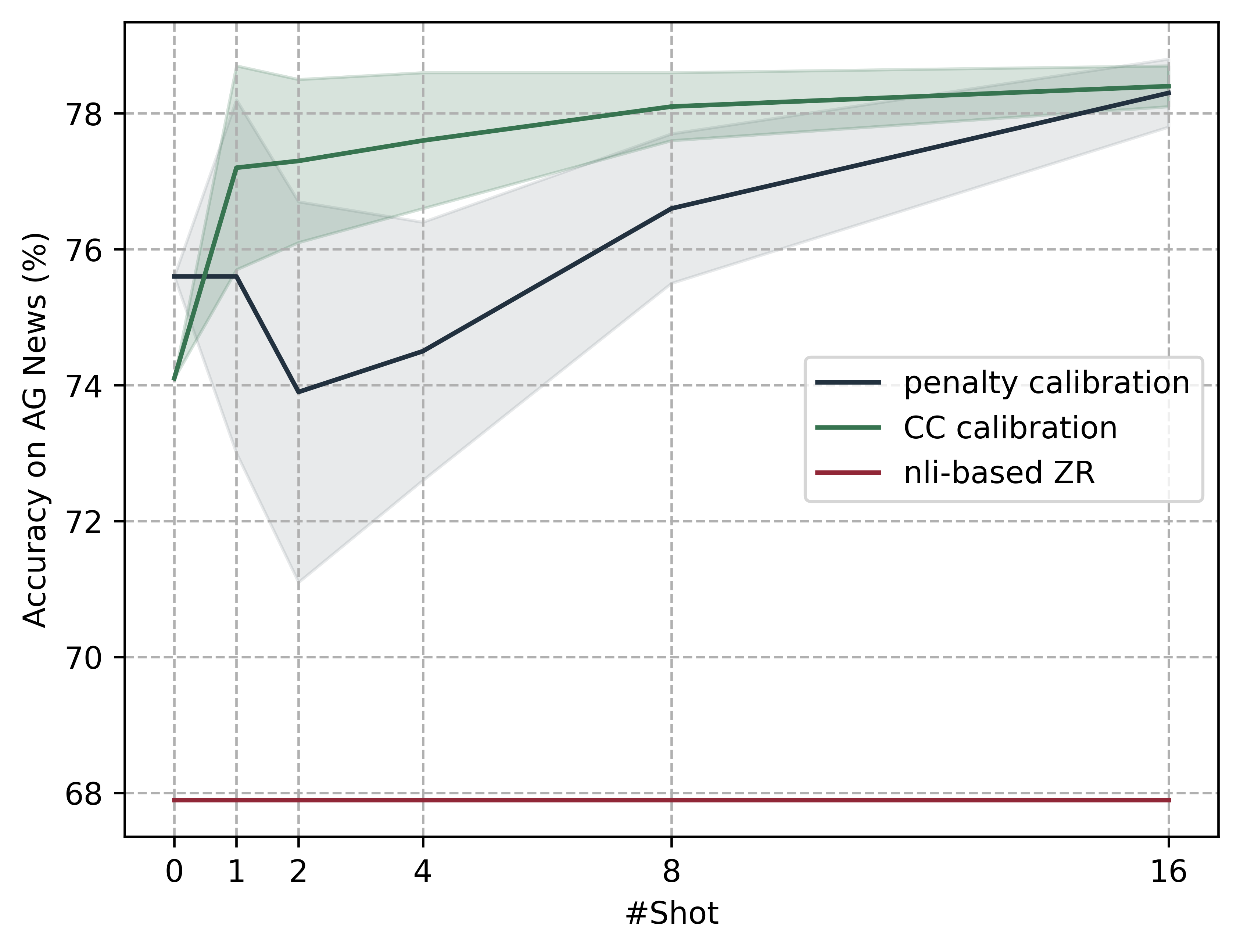}
\endminipage\hfill}
\subfigure[NLI]{
\minipage{.5\textwidth}
  \includegraphics[width=.95\linewidth]{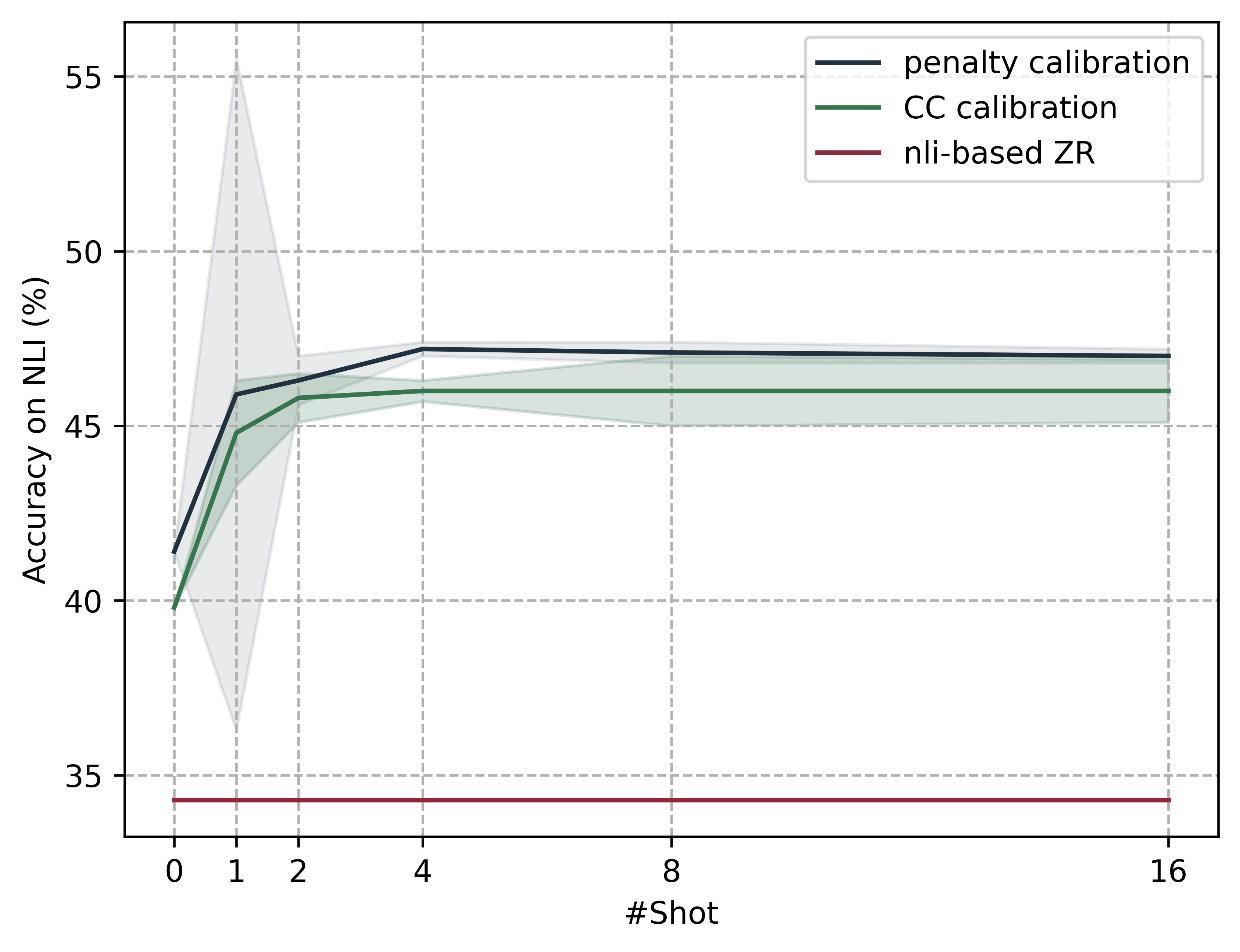}
\endminipage\hfill}
	\caption{Performance and variation of few-shot calibration on the \texttt{RoBERTa} model.}
    \figlabel{plot}
\end{figure}

\subsection{Results on Monolingual Encoders}

\subsubsection{Zero-shot calibration} 
We first validate the effectiveness of the various calibration methods on monolingual encoders. \tabref{results_mono} shows the results of zero-shot calibration, where we directly calculate the calibrated probabilities without using additional training samples.
We report accuracies for evenly distributed datasets and F1 scores for imbalanced datasets.
Compared to the uncalibrated baseline systems, we obtain improvements across most of the tasks, except for the \textit{CC} method combined with the $\texttt{RoBERTa}$ model. In this specific case, the average performance worsens compared to the \textit{no calibration} baseline due to outlier performance observed in several tasks, such as Yahoo and CoLA.

\subsubsection{Adding few-shot samples further boosts the performance} 
As the formulas in \tabref{calibration} show, $\textit{PMI}_{\textit{DC}}$ and \textit{CBM} directly modify the probabilities without introducing additional parameters, while \textit{CC} and \textit{Penalty} use specific calibration parameters, which are trainable. In zero-shot calibration, these parameters are initialized by prior probabilities without being updated. We will now make use of the trainability of parameters in \textit{CC} and \textit{Penalty} to investigate if applying few-shot training to calibration parameters further improves the performance.


\tabref{few_shot_results} shows the results of few-shot calibration. 
We observe that training the calibration parameters on just a few samples further enhances the performance of the calibrated systems.
Compared to zero-shot calibration, few-shot calibration achieves better performance in most cases.
We also  compare calibration methods in few-shot scenarios with the NLI-based zero-shot classification baseline proposed by~\citet{yin-etal-2019-benchmarking}. 
Details of the baseline setting and the few-shot training process are described in Appendices \ref{baseline} and \ref{few_shot_training}.

\figref{plot} shows the few-shot calibration results of the \texttt{RoBERTa} model on the AG News and NLI tasks. 
Prior research \citep{zhao-schutze-2021-discrete} showed that few-shot learning can be unstable due to the randomness. However, as \figref{plot} shows, the variation in performance diminishes obviously as the number of shots increases. Our experimental results indicate that few-shot calibration not only enhances the performance but also increases the steadiness.

\subsection{Results on Multilingual Encoders}

\tabref{multilingual} shows our experimental results on multilingual datasets, indicating that calibration methods are also effective for multilingual encoders.

Our experiments cover a large range of languages considering both language availability, i.e., if or how much language data exists in the pretraining corpus, and language diversity, i.e., to which language family a language belongs. Specifically, for Amazon-S, XNLI and PAWS-X, we use the original test sets, mainly containing high-resource languages. 
In the multilingual AG News task, we include many low-resource and unseen languages by generating parallel multilingual test sets using machine translation techniques.
Recent research by \citet{hu2020xtreme} and \citet{liu-etal-2022-mulzdg} shows that automatically translated test sets are useful for measuring cross-lingual performance. Hence, we adopt their methodology and expand the language coverage of the AG News dataset to 25. The list of languages can be found in Appendix \ref{detail_results}.

The results on multilingual \texttt{BERT} and \texttt{XLM-R} show that all four calibration methods improve the multilingual performance averaged across all tasks. 
For both models, \textit{CBM} always emerges as the top-performing approach. Different from other approaches predicting the label with one input by another, \textit{CBM} is the only method which leverages the test set (without labels) to adjust the calibration parameters. This could account for the substantial advantage of \textit{CBM} over the others in terms of the performance.

 \begin{table*}[h]
\centering
\footnotesize
\begin{tabular}{l|c|c|c|c|c} 
& \textbf{AG News}               & \textbf{Amazon-S}              & \textbf{XNLI}                  & \textbf{PAWS-X}                & \textbf{Avg.}                   \\ 
\hline
\multicolumn{1}{l|}{$\texttt{mBERT}_{\texttt{Base}}$}    & \multicolumn{1}{l|}{} & \multicolumn{1}{l|}{} & \multicolumn{1}{l|}{} & \multicolumn{1}{l|}{} & \multicolumn{1}{l}{}  \\ 
\hspace{1em} + \textit{no calib.}                                               & 32.8                  & 20.5                  & 33.6                  & 33.9                  & 30.2                   \\ 

\hspace{1em} + $\textit{PMI}_{\textit{DC}}$                                                & 48.8                  & 22.5                  & 33.6                  & 44.4                  & 37.3                   \\ 

\hspace{1em} + \textit{CBM}                                                    & 53.8                  & \textbf{25.1}                  & 34.9                  & \textbf{49.2}                  & \textbf{40.8}                   \\ 
\hspace{1em} + \textit{CC (max)}                                               & 53.9                  & 23.9                  & 35.1                  & 44.8                  & 39.4                   \\ 

\hspace{1em} + \textit{Penalty (max)}                                          & \textbf{54.6}                  & 23.8                  & \textbf{35.3}                  & 47.1                  & 40.2                   \\ 
\hline
\multicolumn{1}{l|}{$\texttt{XLM-R}_{\texttt{Base}}$} & \multicolumn{1}{l|}{} & \multicolumn{1}{l|}{} & \multicolumn{1}{l|}{} & \multicolumn{1}{l|}{} &                        \\ 

\hspace{1em} + \textit{no calib.}                                               & 45.4                  & 21.9                  & 35.0                  & 31.7                  & 33.5                   \\ 

\hspace{1em} + $\textit{PMI}_{\textit{DC}}$                                                & 59.8                  & 23.0                  & 33.6                  & 37.8                  & 38.6                   \\ 

\hspace{1em} + \textit{CBM}                                                    & \textbf{63.3}                  & \textbf{28.9}                  & \textbf{37.8}                  & \textbf{46.3}                  & \textbf{44.1}                   \\ 

\hspace{1em} + \textit{CC (max)}                                               & 59.6                  & 23.7                  & 35.3                  & 43.7                  & 40.6                   \\ 

\hspace{1em} + \textit{Penalty (max)}                                          & 57.5                  & 23.6                  & 35.8                  & 43.4                  & 40.1                   \\
\hline
\end{tabular}
\caption{Results of calibration methods on multilingual datasets. We report the best results for \textit{CC} and \textit{Penalty} in different few-shot settings.}
\tablabel{multilingual}
\end{table*}

\subsection{Multilingual Analysis}
Now we analyze how different language properties correlate with the
performance of multilingual BERT on the AG News task.

\subsubsection{Language Accessibility} 
We first group the evaluation languages into low-resource languages, unseen languages, and languages with unseen scripts to determine the influence of language accessibility. Low-resource languages are languages which are contained in the pretraining corpus, but only account for a small amount of it. Unseen languages do not occur in the pretraining, thus the multilingual encoder has never seen them. The hardest case involves languages with unseen scripts, where the model has not even encountered the characters of the language. 
However, our test set contains no languages with completely unseen scripts because machine translation frequently generates code-switched data.
\figref{analysis} (a) shows that low-resource languages perform generally better than the other two types of unseen languages, indicating that the multilingual encoder's access to languages in the pretraining is crucial for the performance enhancement via calibration.

\subsubsection{Language Diversity} We further group the languages according to their phylogenetic relationships, i.e., from which language family they are. We analyze the language families containing at least 3 languages. The box plots in \figref{analysis} (b) reveal that the impact of calibrating multilingual encoders varies across different language groups. Specifically, we observe that Indo-European and Dravidian languages tend to benefit more from calibration than Austronesian and Niger-Congo languages.

\begin{figure}[!t]
\subfigure[Language Accessibility]{
\minipage{.5\textwidth}
  \includegraphics[width=.95\linewidth]{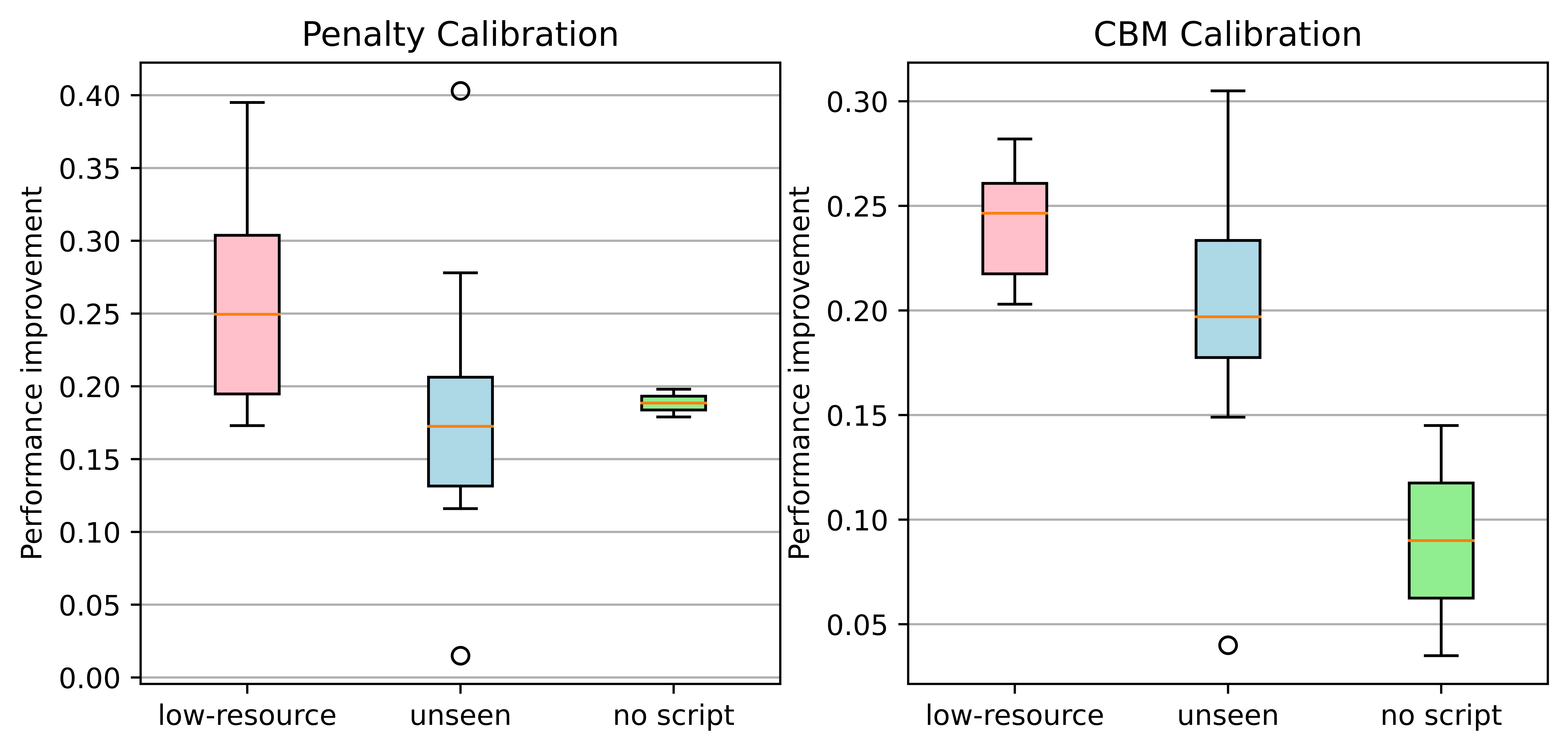}
\endminipage\hfill}
\subfigure[Language Diversity]{
\minipage{.5\textwidth}
  \includegraphics[width=.95\linewidth]{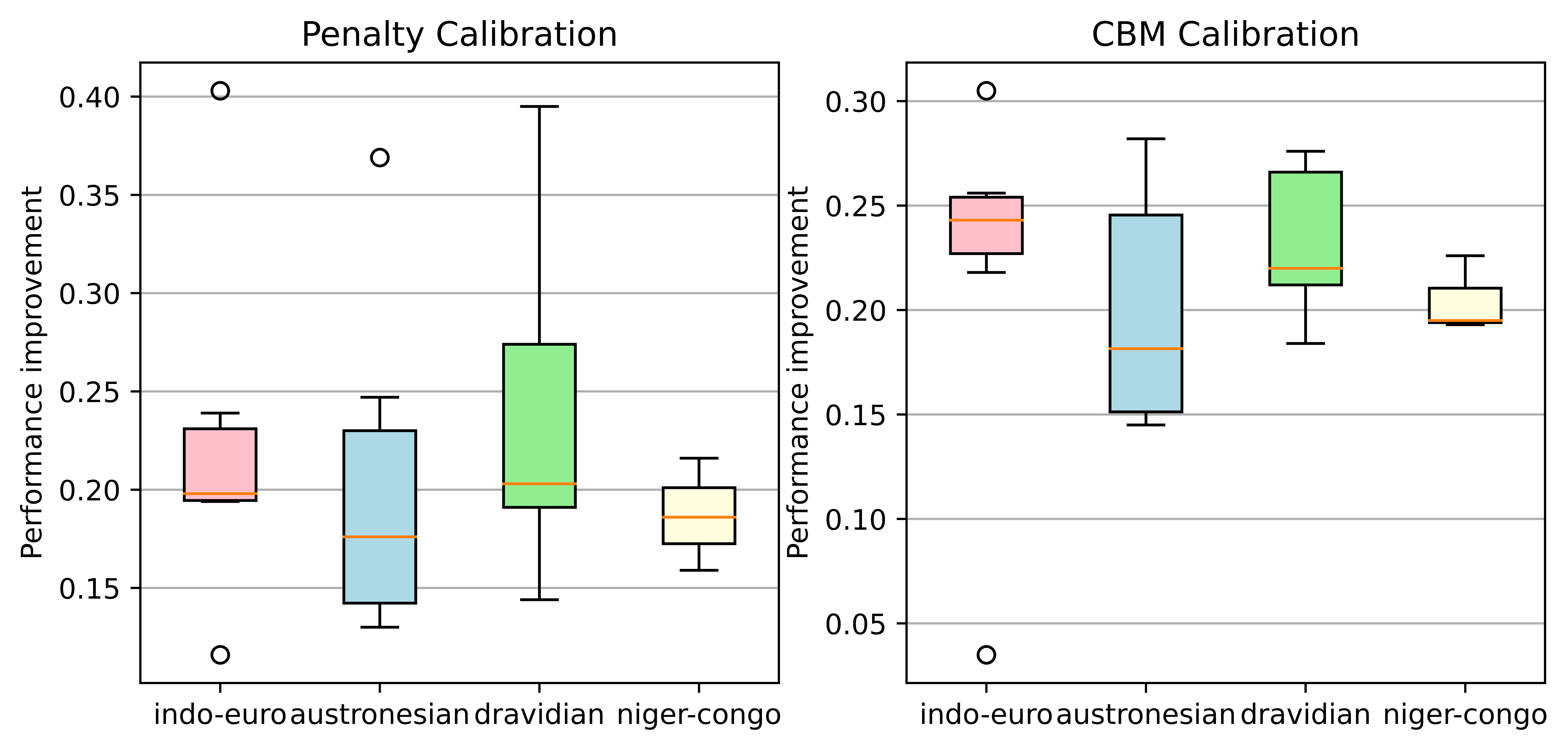}
\endminipage\hfill}
	\caption{Performance Improvement of multilingual BERT with two calibration methods.}
    \figlabel{analysis}
\end{figure}

\paragraph{}This discrepancy suggests that the effectiveness of calibration techniques can be influenced by the language accessibility of multilingual encoders and the linguistic characteristics of language families.

\section{Conclusion} 
In conclusion, our work focuses on boosting the zero-shot learning performance of multilingual encoders in language understanding tasks through probability calibration techniques. We address the bias issue in the mask token prediction of label words by introducing various calibration techniques that modify the probabilities of these words. 
We first test the efficacy of different calibration methods in monolingual encoders. We also prove that with a minimal number of training examples, the calibrated probabilities yield further enhancements compared to the zero-shot calibration method.
Our experiments on multilingual encoders demonstrate that all calibration methods bring a performance improvement across various tasks.

\section*{Limitations}
We propose a simple yet effective calibration method to enhance the zero-shot performance for monolingual and multilingual encoders. 
While our work shows the effectiveness of calibration for enhancing the prediction with multilingual tasks, it is important to note that our research is primarily focused on classification tasks with multilingual encoders. As a result, our findings and proposed methods may not directly translate to generation tasks, such as question answering (QA), which involve the use of generative multilingual models. Future investigations should explore the application of our calibration methods on generation tasks and evaluate their effectiveness in enhancing the performance of generative multilingual models. This extension could provide valuable insights into the potential benefits and limitations of our approaches across a broader range of NLP tasks.

\section*{Ethics Statement}
This research was conducted in accordance with the ACM Code of Ethics. 
All the datasets that we use are publicly available. 
We report only aggregated results in the main paper.
We do not share any Personally Identifiable Data in this paper.

\section*{Acknowledgements}
We extend our sincere gratitude to the anonymous reviewers for their invaluable contributions and constructive feedback that have greatly enriched the quality and scope of this paper. This work was supported by Munich Center for Machine Learning (MCML) and China Scholarship Council (CSC).

\bibliography{anthology,custom}
\bibliographystyle{acl_natbib}

\appendix

\section{Experimental Details}
\label{details}
 This section provides a comprehensive overview of our experimental setup, including hyperparameters, prompt templates that we use in our experiments, and the baselines.

\subsection{Hyperparameters}

To ensure experimental reproducibility, we present the hyperparameter settings used in our study in \tabref{hyperparameters}.

\begin{table}[h!]
\centering
\scriptsize
\begin{tabular}{|l|l|} 
\hline
\textbf{Hyperparameter}          & \multicolumn{1}{l|}{\textbf{Value}}                                                                              \\ 
\hline
Evaluation batch size   & 8                                                                                                       \\ 
\hline
Learning rate           & 1e-4                                                                                                \\ 
\hline
Random seeds            & $\text{\{42, 421, 512, 1213, 1234\}}$  \\ 
\cline{1-1}\cline{2-2}
Maximal sequence length & 128                                                                                                     \\ 
\hline
Few-shot numbers        & $\text{\{1, 2, 4, 8, 16\}}$                                                                 \\ 
\hline
GPU type                & \multicolumn{1}{l|}{NVIDIA GeForce GTX 1080 Ti}                                    \\ 
\cline{1-1}\cline{2-2}
Number of GPU           & 8                                                                                                       \\
\hline
\end{tabular}
\caption{Overview of hyperparameters.}
\tablabel{hyperparameters}
\end{table}

\subsection{Prompt Engineering}
\label{prompt_engi}
We select a set of prompt templates for the tasks through our preliminary experiments. \tabref{prompts} shows the prompt templates and the label words used in our experiment.
\begin{table*}
    \centering
    \scriptsize
    \begin{tabular}{l|l|l}
      \textbf{Task}   &  \textbf{Prompt template} & \textbf{Label words} \\
      \hline
        Ag News & $\texttt{mask} \text{ News: }\texttt{[X]}$ & `World', `Sports', `Business', `Tech'\\
        Amazon-P & $\texttt{[X]} \text{. All in all, it was }\texttt{mask} \text{.}$ & `bad', `good'\\  
        Amazon-P & $\texttt{[X]} \text{. All in all, it was }\texttt{mask} \text{.}$ & `terrible', `bad', `ok', `good', `great'\\
        XNLI & $\texttt{[X]} \text{? }\texttt{mask} \text{, }\texttt{[Y]}$ & `Yes', `Maybe', `No'\\
        Yahoo & $\texttt{mask} \text{ Question: }\texttt{[X] [Y]}$ & `Society', `Science', `Health', `Education', $\cdots$ \\       PAWS-X & $\texttt{[X]} \text{ . }\texttt{mask} \texttt{[ Y]}$ & `Wrong', `Right'\\  
         CoLA & $\texttt{[X]} \text{ . It is linguistially }\texttt{mask} \text{.}$ & `wrong', `right'\\  
        MRPC & $\texttt{[X]} \text{? }\texttt{mask} \text{, }\texttt{[Y]}$ & `Wrong', `Right'\\
        QQP & $\text{Question 1: } \texttt{[X]} \text{ Question 2:  } \texttt{[Y]} \text{ Question 1 and Question 2 are } \texttt{mask} $ & `different', `same'\\
        RTE & $\texttt{[X]} \text{? }\texttt{mask} \text{, }\texttt{[Y]}$ & `Wrong', `Right'\\
        WNLI & $\texttt{[X]} \text{? }\texttt{mask} \text{, }\texttt{[Y]}$ & `Wrong', `Right'\\
        \bottomrule
    \end{tabular}
    \caption{Overview of prompt templates.}
    \tablabel{prompts}
\end{table*}

\subsection{Baseline}
\label{baseline}
To establish a baseline, we initially conduct experiments without employing any calibration methods. Subsequently, we introduce four calibration methods individually and evaluate their impact on the performance. 
Besides, we compare our calibration methods with an NLI-based zero-shot classification baseline proposed by \citet{yin-etal-2019-benchmarking}, where they first finetune a pretrained language model on the MNLI dataset, then they reformulate common classification tasks to an NLI task format. The input sample is regarded as the premise, while the label serves as the hypothesis. The zero-shot classification is performed by directly comparing the probabilities of predicting \texttt{entailment} for all input-label pairs.
For this baseline, we finetune a \texttt{BERT} model and a \texttt{RoBERTa} model on the MNLI task.

\section{Few-Shot Training of Calibration Parameters}
\label{few_shot_training}
Algorithm \ref{algo1} presents the process of few-shot training of penalty calibration used in our few-shot investigation. 

\begin{algorithm}
    \SetAlgoLined
    \KwIn{set of few-shot training samples $\textbf{\textit{D}}$, initial calibration parameter vector $\textbf{\textit{p}}_0$, number of epochs $E$, learning rate $\eta$}
    \KwOut{Trained parameters $\textbf{\textit{p}}$}
    \BlankLine
    Initialize $\textbf{\textit{p}} \leftarrow \textbf{\textit{p}}_0$\;
    
    \For{epoch \textbf{in} $1,2,\cdots,E$}{
        \ForEach{$(x, y)$ \textbf{in} $\textbf{D}$}{
        $\textbf{\textit{l}} \leftarrow get\_probs(x)$\;
        
        $\textbf{\textit{l}}_y\leftarrow \textbf{\textit{l}} - \textbf{\textit{p}}$
        \CommentSty{\# calibration}\;

        $\hat{y}\leftarrow argmax_y(\textbf{\textit{l}}_y)$\;

        \If{$y\neq \hat{y}$}{
        $\textbf{\textit{p}}_{t}[\hat{y}]\leftarrow \textbf{\textit{p}}_{t}[\hat{y}] +\eta $\;

        $\textbf{\textit{p}}_{t}[y]\leftarrow \textbf{\textit{p}}_{t}[y] -\eta $\;
        }
        }
    }
    \caption{Few-Shot Training of Penalty Calibration}
    \label{algo1}
\end{algorithm}

\section{Detailed Results}
\label{detail_results}

Detailed results of the experiments in the main text can be found in this section. \tabref{ag_news_all} shows the complete results of mBERT on the multilingual AG News dataset across all 25 languages. \tabref{langs} provides an overview of languages covered by the multilingual AG News dataset.

\begin{table*}[h]
\centering
\begin{tabular}{c|l|l|l} 
\hline
\textbf{Code} & \textbf{Languages}      & \textbf{Language Accessibility} & \textbf{Language Family}                                   \\ 
\hline
af   & Afrikaans      & Low-resource           & Indo-European                                     \\ 
co   & Corsican       & Unseen languages       & Indo-European                                     \\ 
eo   & Esperanto      & Unseen languages       & Artificial                                        \\ 
haw  & Hawaiian       & Unseen languages       & Austronesian                               \\ 
hmn  & Hmong          & Unseen languages       & Sino-Tibetan                                      \\ 
ht   & Haitian Creole & Low-resource           & Indo-European                                     \\ 
ig   & Igbo           & Unseen languages       & Niger-Congo                                       \\ 
jw   & Javanese       & Low-resource           & Austronesian                                      \\ 
km   & Khmer          & Unseen script          & Austronesian                                      \\ 
mi   & Maori          & Low-resource           & Austronesian                                      \\ 
mn   & Mongolian      & Low-resource           & mongolian                                         \\ 
mt   & Maltese        & Unseen languages       & Afro-Asiatic  \\ 

my   & Burmese        & Low-resource           & Sino-Tibetan                                      \\ 
ny   & Chichewa       & Unseen languages       & Niger-Congo                                       \\ 
or   & Odia           & Unseen script          & Indo-European                                     \\ 
sm   & Samoan         & Unseen languages       & Austronesian                                      \\ 
sn   & Shona          & Unseen languages       & Dravadian                                         \\ 
st   & Sesotho        & Unseen languages       & Dravadian                                         \\ 
sw   & Swahili        & Low-resource           & Dravadian                                         \\ 
ta   & Tagalog        & Low-resource           & Austronesian                                      \\ 
te   & Telugu         & Low-resource           & Dravadian                                         \\ 
tl   & Tamil          & Low-resource           & Dravadian                                         \\ 
ug   & Uighur         & Unseen languages       & Turkic                                            \\ 
ur   & Urdu           & Low-resource           & Indo-European                                     \\ 
uz   & Uzbek          & Low-resource           & Turkic                                            \\ 
zu   & Zulu           & Unseen languages       & Niger-Congo                                       \\
\hline
\end{tabular}
\caption{Overview of languages covered by the multilingual AG News dataset.}
\tablabel{langs}
\end{table*}

\begin{table*}
\centering
\footnotesize
\begin{tabular}{l|c|c|c|c|c|c|c|c|c|c|c|c|c|c} 
\toprule
          & af   & co   & en   & eo   & haw  & hmn  & ht   & ig   & jw   & km   & mi   & mn   & mt   & my    \\ 
\hline
No calib. & 40.4 & 32.6 & 47.3 & 31.9 & 27.1 & 30.9 & 35.7 & 30.2 & 38.0 & 33.3 & 29.0 & 32.0 & 29.9 & 33.8  \\ 

Penalty   & 64.3 & 44.2 & 69.6 & 72.3 & 40.1 & 49.6 & 55.2 & 48.8 & 62.6 & 51.2 & 46.3 & 62.2 & 57.6 & 64.7  \\ 

CBM       & 64.7 & 58.3 & 69.1 & 62.4 & 42.0 & 50.8 & 60.9 & 49.6 & 63.9 & 47.8 & 49.5 & 53.0 & 57.2 & 54.1  \\ 

CC        & 65.6 & 59.7 & 67.8 & 68.0 & 43.4 & 49.7 & 65.2 & 52.4 & 66.4 & 41.4 & 51.2 & 55.4 & 57.4 & 51.7  \\ 

$\text{PMI}_\text{DC}$       & 60.2 & 35.3 & 60.0 & 61.7 & 35.9 & 33.5 & 33.5 & 49.2 & 61.5 & 42.2 & 49.6 & 54.7 & 61.1 & 47.6  \\ 
\midrule
          & ny   & or   & sm   & sn   & st   & sw   & ta   & te   & tl   & ug   & ur   & uz   & zu   & avg.  \\ 
\hline
No calib. & 29.8 & 25.4 & 30.3 & 32.2 & 30.4 & 33.4 & 28.8 & 32.5 & 42.6 & 25.5 & 33.2 & 33.9 & 34.5 & 32.8  \\ 

Penalty   & 51.4 & 45.2 & 43.5 & 52.4 & 44.8 & 72.9 & 65.6 & 59.9 & 61.7 & 27.0 & 52.6 & 59.1 & 50.3 & 54.6  \\ 

CBM       & 52.4 & 28.9 & 46.1 & 53.4 & 48.8 & 59.9 & 57.0 & 60.0 & 64.6 & 29.5 & 56.8 & 58.9 & 53.7 & 53.8  \\ 

CC        & 51.2 & 28.7 & 47.5 & 52.5 & 49.1 & 64.1 & 56.5 & 52.4 & 62.6 & 27.9 & 53.1 & 60.3 & 49.6 & 53.7  \\ 

$\text{PMI}_\text{DC}$       & 50.2 & 28.6 & 43.9 & 50.9 & 44.6 & 61.6 & 50.1 & 43.6 & 66.1 & 29.3 & 55.0 & 56.4 & 51.3 & 48.8  \\
\bottomrule
\end{tabular}
\caption{Results of mBERT on the multilingual AG News dataset across all languages.}
\tablabel{ag_news_all}
\end{table*}

\end{document}